\documentclass[runningheads]{llncs}

\usepackage[T1]{fontenc}
\usepackage{graphicx}
\usepackage[table]{xcolor}

\usepackage{soul}
\usepackage{url}

\definecolor{hypercolor}{rgb}{0.0, 0.18, 0.39}

\usepackage{hyperref}
\hypersetup{
colorlinks=true,
linkcolor=hypercolor,
citecolor=hypercolor,
urlcolor=hypercolor,
pdftitle={ASPEn -- Answer Set Programming Energised!},
pdfauthor={Anonymous},
pdfsubject={ASPEn -- Answer Set Programming Energised!},
pdfcreator={},
pdfproducer={}
}

\usepackage[utf8]{inputenc}
\usepackage[T1]{fontenc}
\usepackage[small]{caption}
\usepackage{graphicx}
\usepackage{amsmath}
\usepackage{booktabs}
\usepackage{algorithm}
\usepackage{algorithmic}
\usepackage{amsfonts}

\usepackage{subcaption}

\usepackage[frozencache]{minted}
\setminted{fontsize=\scriptsize}

\newcommand{\ASPenergised}{{ASPEn}}

\newcommand{\mintedBGColor}{blue!5!white}

\newcommand{\myExample}[2]{
\begin{example}
{#2}
\end{example}
\label{#1}
}

\usepackage{lipsum}

\definecolor{headerblue}{RGB}{220, 232, 246}
\definecolor{labelblue}{RGB}{235, 243, 252}

\begin{document}

\tolerance=1000 \emergencystretch=2em

%
\title{{Answer Set Programming Energised! }\\[4pt]{\large End-to-End Neurosymbolic Reasoning and Learning with\\[-3pt]ASP and Energy Based Models}}
\titlerunning{ASPEn -- \emph{Answer Set Programming Energised}! }
%

%

\author{Jakob Suchan\inst{1,3} \and
Julius Monsen\inst{2,3} \and
Salim Baloch\inst{1,3} \and
Mehul Bhatt\inst{2,3}}

\authorrunning{Suchan et al.}

\institute{
Constructor University Bremen, Bremen, Germany\\
\email{jsuchan@constructor.university} \and
Örebro University, Örebro, Sweden\\
\email{info@codesign-lab.org} \and
CoDesign Lab > Cognitive Vision\\
\href{https://codesign-lab.org/cognitive-vision/}{codesign-lab.org/cognitive-vision}}

\maketitle              
%
%

\begin{abstract}
We present a general neurosymbolic reasoning and learning methodology based on a modular integration of \emph{answer set programming} with an \emph{energy based model} substrate. Key contributions are: {\bfseries(1)} supporting joint optimisation in the continuous latent space through explicit ASP-based declarative semantics fully incorporating background knowledge, constraints, non-monotonic inference; and {\bfseries(2)} advancing recent works at the interface of answer sets, probabilistic logic, and answer set modulo theories by providing a generalised model and practical platform for ASP-centric robust, end-to-end training for applications in \emph{dynamic domains} (e.g., involving perception  and interaction). We provide a practical implementation, and demonstrate basic use and application (with {\footnotesize MNIST}), and evaluate with the visual question-answering benchmark Clevr and the multi-object tracking benchmark {\footnotesize MOT}.
\end{abstract}

%
%
%
%

\section{Motivation}

The integration of machine learning and reasoning remains a challenge and an area of high interest in AI research \cite{nesy3rd-wave2023}. Addressing this is particularly needed for  real-world problems  --e.g., in embodied perception, control, decision-making-- involving interactive dynamics, uncertainty \& partial observability, abnormalities etc \cite{Bhatt21_Artificial_Visual_Intelligence,DBLP:conf/kr/SuchanBM25}. Towards this, the broader purpose of this research is on high-level, semantically-guided learning of high-dimensional sensory structure keeping in mind trustworthy design. i.e., explainability, interpretability, formal verification-- aimed at ethico-legal compliance vis-a-vis emerging AI regulation \cite{HLEG-AI-2019,Regulation-2021,Nordic-AI-Regulation-2024}. The main scientific aim is to advance  systematic, robust methodological developments aimed at integrating the declaratively modelled semantic structure of problem spaces with quantitative optimisation and learning.

%


\medskip

\textbf{Declarative Neurosymbolism, End-to-End}.\quad Neurosymbolic integrations from the viewpoint of KR and ML research have gained traction in recent years. Particularly relevant to the scope addressed here are declarative methodology centric integrations involving stable model semantics rooted Answer Set Programming (ASP) \cite{ASP-Glance-2011}, and ASP derivatives such as specialised Answer Set Modulo Theories (ASPMT) \cite{functionalASP-2013,TPLP-ASPMTQS} and Probabilistic ASP \cite{lpmln}. Most recently, an active line of work building upon such fundamental perspectives pertains integration of KR/ASP and (deep learning driven) computer vision aimed at realising neurosymbolic visual commonsense  \cite{DBLP:conf/kr/SuchanBM25,sqa-suchan16,outofsight-ijcai2019,Yang2020_NeurASP,AIJ2021-Suchan,pmlr-v284-eiter25a,PADALKAR_GUPTA_2025}. Here,  the use of logic and answer set programming stands out as a sustained line of inquiry from viewpoints such as  (semantic) visual-question answering \cite{sqa-suchan16,pmlr-v284-eiter25a,PADALKAR_GUPTA_2025}, and non-monotonic visual abduction for (neurosymbolic) spatio-temporal belief maintenance in dynamic domains \cite{AIJ2021-Suchan}, and ``in-the-wild'' neurosymbolic reasoning about embodied interaction \cite{Suchan2025_KR}. Differences in the technical framing and supported computational capabilities in these works notwithstanding, the general underlying motivation remains unified: integrating neurally-driven visual processing capabilities (to extract geometric scene elements and visual features from imagery) with high-level, expressive conceptual commonsense knowledge with the aim to support  neurosymbolic interpretation of either static and/or dynamic stimuli. 

With these methodologies, a caveat is that they preserve a conceptual and computational separation between symbolic inference and subsymbolic learning: neural outputs are injected into a logical program, reasoning is performed over discrete abstractions, and learning is driven primarily by task-level losses {instead of} the full declarative semantics of the reasoning mechanism (i.e., full expressivity afforded by a non-monotonic framework such as ASP). This separation becomes a fundamental limitation in non-trivial, real-world dynamic domains, e.g., in problems such as real-time vision where truly integrated reasoning and learning requires the joint resolution of aspects such as perceptual uncertainty, dynamic spatial-temporal consistency, and non-monotonic commonsense reasoning for belief revision over extended time horizons. For truly \emph{integrated end-to-end reasoning and learning}, the inference mechanism cannot merely be a filtering or validation step, but should be an integral part of the learning process itself. What is needed, therefore, is a neurosymbolic framework where symbolic semantics, optimisation, and learning operate as  modular yet tightly integrated and mutually influencing computational substrates.

\medskip

\textbf{ASPEn -- \emph{ASP Energised}!}\quad We propose a general, declarative neurosymbolic reasoning and learning methodology based on an integration of  ASP with Energy-Based Models (EBM). The core operational mechanism involves interpreting ASP-based world models as structured  embeddings in a continuous optimisation landscape. In this setting, ASP-based stable models are no longer treated as isolated symbolic artefacts, but function as preferred configurations in a joint discrete–continuous space, obtained through semantically-guided energy minimisation. Previous ASP and logic programming based neurosymbolic methods involving weighted rule satisfaction, probabilistic choice over stable models, constraint consistency in answer set modulo theories, and likelihood-based neural predictions all induce preferences over interpretations. Our approach, essentially a generalisation with an energy-based formulation, makes this preference structure explicit and operational; i.e., logical constraints, defaults, abductive explanations, as well as specialised domains for aspects such as space and motion directly \emph{shape the energy landscape over candidate world models}. Importantly, this formulation supports bidirectional interaction, e.g., non-monotonic reasoning not only constrains inference, but also generates structured learning signals that influence the optimisation of the low-level energy landscape itself. In our approach,  ASP’s native optimisation constructs enable preference-based selection among competing interpretations, a capability that aligns naturally with energy minimisation. We demonstrate the proposed method through a practical implementation and its empirical evaluation spanning both controlled as well as complex dynamic settings. Addressing community benchmarks, we also perform problem-domain specific evaluation focussing on compositional visual reasoning (\emph{static case}; {\small\bfseries CLEVR} \cite{Johnson17}), and
 multi-object tracking (\emph{dynamic case}; {\small\bfseries MOT}) involving (commonsense) neurosymbolic reasoning and learning about space and motion. These experiments illustrate that the proposed methodology scales beyond {\small\bfseries MNIST} \cite{lecun-mnisthandwrittendigit-2010} examples, supports rich declarative modelling, and yields robust performance (under real-world challenges such as noise, ambiguity, partial observability).

\section{Foundational Preliminaries}\label{sec:prelim}
We present the necessary foundational preliminaries relevant to  declarative non-monotonic reasoning and low-level optimisation and learning involving Answer Set Programming and Energy Based Models respectively:

\subsection{Answer Set Programming (ASP)}\label{sec:asp-preliminaries}

Answer Set Programming is an established and widely applied foundational declarative language and robust computing methodology for a range of (non-monotonic) knowledge representation and reasoning tasks \cite{ASP-SI-2018,Gebser2012-ASP,Gebser2014-Clingo,ASP-Glance-2011,ASP-books/sp/Lifschitz19}. Rooted in the \emph{stable model semantics} of logic programs, reasoning in ASP is equivalent to model-theoretic derivation and selection/optimisation of  interpretations, i.e., stable models or answer sets, under non-monotonic assumptions and support for defaults, exceptions, indirect effects, preferences. In essence, ASP supports declarative selection among alternative models via minimisation and optimisation, making it particularly suited to diverse forms of commonsense reasoning, e.g., abductive explanation in the presence of incomplete information \cite{AIJ2021-Suchan}. 

\medskip

\textbf{Basic ASP Semantics}.\quad An answer set program is a finite set of rules of the form $h \;\! \leftarrow \;\! b_1,\dots,b_m,\; \text{not } c_1,\dots,\text{not } c_n$, where $h$, $b_i$, and $c_j$ are (possibly strongly negated) atoms, and $\text{not}$ denotes \emph{default negation}. Intuitively, the rule states that $h$ is derivable if all positive body literals $b_i$ hold and none of the
negative literals $c_j$ can be derived. 

\smallskip

Let $\Pi$ be a ground ASP program and $I$ an interpretation. The
{Gelfond--Lifschitz reduct} of $\Pi$ with respect to $I$ is obtained by:  {\small\textbf{(1)}} discarding every rule $r \in \Pi$ whose body contains a default-negated literal $\text{not } c$ with $c \in I$; and {\small\textbf{(2)}} removing all remaining default-negated literals from rule bodies. 

The resulting program $\Pi^I$ is a positive (negation-free) logic program. An interpretation $I$ is a {stable model} or an {answer set}) of $\Pi$ if $I$ is a \emph{minimal model} (under set inclusion) of the reduct of $\Pi$ with respect to $I$. We denote a stable model of an ASP program as $\mathbf{SM}$[$\Pi$]. Stable models constitute self-supporting interpretations that satisfy all rules under a non-monotonic reading. It is important to note that in this setting, reasoning is model-selective rather than proof-oriented, and that multiple stable models may exist, each corresponding to a distinct, coherent world model.





\subsection{Energy-Based Model (EBM)}\label{sec:ebm-preliminaries}

Energy-Based Models (EBMs) formulate learning and inference as optimisation over a scalar-valued energy function that scores the compatibility of structured configurations \cite{lecun2006tutorial}.  An EBM is defined by an \emph{energy function} $\mathbf{E_\theta(x)} : \mathcal{X} \rightarrow \mathbb{R}$, where $\mathcal{X}$ denotes the space of possible configurations (e.g., data, latent variables, or structured hypotheses), and $\theta$ denotes a set of
learnable parameters. Lower energy values correspond to more compatible or
preferred configurations. Optimisation in EBM is formulated as:
$x^* = \arg\min_{x \in \mathcal{X}} E_\theta(x)$, selecting configurations that best satisfy learned and imposed constraints.
Unlike probabilistic models, EBMs do not require normalised probability
distributions; relative energy differences suffice to define preferences among
configurations.

Learning in EBMs proceeds by shaping the energy landscape so that observed or
desired configurations attain lower energy than implausible ones. Given a set
of target configurations $\mathcal{D}$, training typically minimises a loss of
the form 
$ \mathcal{L}(\theta) = E_\theta(x^+) - E_\theta(x^-)$, where $x^+$ denotes data-consistent (positive) configurations and $x^-$ denotes negative or contrastive configurations obtained via sampling or optimisation. This contrastive formulation supports weak supervision, partial observability, and structured prediction. The energy function is often decomposed additively: $E_\theta(x,y) = \sum_{i} E_{\theta_i}(x,y)$, where each term corresponds to a constraint, interaction, or preference. This supports modular integration of heterogeneous knowledge sources.

A key characteristic of EBMs is that these represent learning and reasoning entirely in terms of compatibility constraints over configurations, subsuming generative and discriminative models as special cases while avoiding commitment to explicit likelihoods or fixed input–output mappings  \cite{lecun2006tutorial}. Multiple, heterogeneous constraints --e.g., originating from perception or domain knowledge-- can be combined additively as energy terms, enabling joint optimisation over modularly integrated hybrid configurations.  EBMs are therefore particularly suited for settings that require integration of continuous representations with symbolic or relational structure in the manner pursued in this research.

\section{{ASP Energised}! Formal Framework and Implementation}
ASPEn is a novel methodology integrating answer set programming and energy-based models for declarative neurosymbolic reasoning and learning in a manner that is end-to-end, and designed for challenging dynamic domains. Building upon recent and emerging works in neurosymbolic ASP (Sec \ref{sec:prelim}, \ref{sec:disc}), we offer a general community-wide platform for a unified pursuit of research, development, evaluation and benchmarking of efforts in ASP-centric initiatives aimed at integrating KR and ML.

\begin{figure*}[t]
\center
\includegraphics[width=\textwidth]{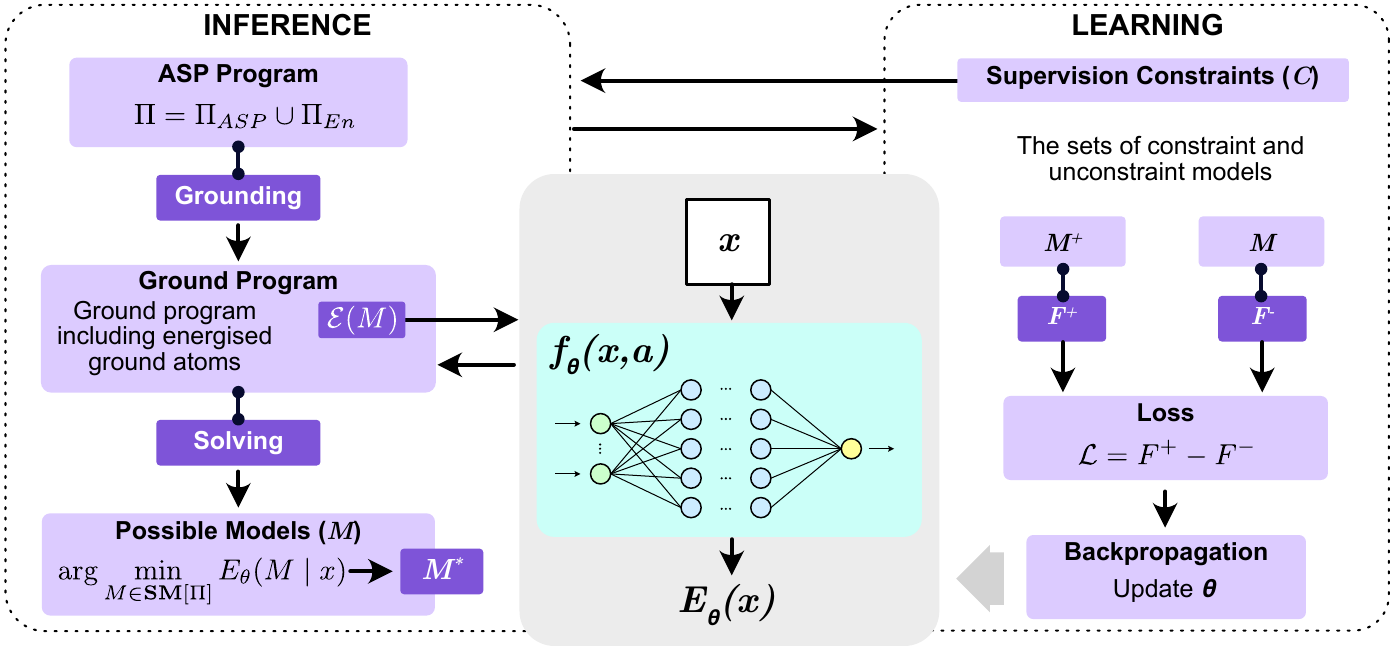}
\caption{\textbf{ASPEn}: End-to-End Inference and Learning Architecture}
\label{fig:ASPEn}
\end{figure*}

\subsection{Syntax and Semantics of \ASPenergised{}}\label{sec:synt-semantics}
An \ASPenergised{} program $\Pi$ is composed of a classical ASP program $\Pi_{ASP}$ and a set of energised atoms $\Pi_{En}$, i.e., $\Pi~\equiv_{def}~\Pi_{ASP}~ \cup~\Pi_{En}$. Again, $\mathbf{SM}$[$\Pi$] denotes the set of stable models (Sec \ref{sec:asp-preliminaries}) of $\Pi$. Below we primarily focus on succinctly and pragmatically highlighting the modelling of $\Pi_{En}$ component while assuming basic familiarity with declarative modelling of ASP programs wrt. the preliminaries in Sec \ref{sec:asp-preliminaries}.

\medskip

\textbf{ASPEn Syntax focussing on $\Pi_{En}$.} \quad Energies are obtained from Energy Based Models (EBMs) referenced by specific energy atoms:



\begin{minted}[bgcolor=\mintedBGColor, tabsize=2]{prolog} 
energise(rule_name([Args]))  :- B.
\end{minted}

Here, the energise predicate couples an ASP rule $rule\_name([Args])$ with an energy function implemented as EBM (Sec \ref{sec:ebm-preliminaries}). An EBM take $n$ tensors as input, specified by the arguments ($Args$) in the energising rule, and outputs the corresponding energy $E$. To put this into perspective, consider a minimal working example:


\myExample{ex-mnist}{{\bfseries\small Digit-value assignment in MNIST}.}


Consider a minimal ASP program that assigns a numeric value (0 - 9) to an image of a handwritten digit, as it is for instance part of the `Addition' task within the MNIST dataset (Sec. \ref{sec:MNIST}).
Such digit assignment may be modelled by the following choice rule:


%
%

\begin{minted}[bgcolor=\mintedBGColor, tabsize=2]{prolog} 
img(img). digit_value(0..9).
\end{minted}
\begin{minted}[bgcolor=\mintedBGColor, tabsize=2]{prolog} 
1{digit(X,V) : digit_value(V)}1 :- img(X).
\end{minted}


To energize the digit assignment in \ASPenergised{}, we can then introduce the following energy term:


\begin{minted}[bgcolor=\mintedBGColor, tabsize=2]{prolog} 
energise(digit(X,V)) :- digit(X,V).
\end{minted}

This term couples the atom $digit(X,V)$ with an energy obtained from an energy function implemented as an EBM that takes the arguments of the atom, i.e., the image and a value, as input and assigns an energy to it. Finding the best assignment of an image to a numeric value is then based on minimising the energy digit-value pair with respect to this assigned energy.

%
%

\smallskip

Building on the basic setup of Example 1, consider now the case of the CLEVR dataset \cite{Johnson17}, which consists of synthetic visual stimuli with abstract geometric entities and corresponding questions:

\myExample{2}{{\bfseries\small Compositional visual question-answering with CLEVR}.}


Similar to the energisation of digit-value assignment in the MNIST task of Example 1, we may now energise multiple atoms representing object properties (e.g., $size$, and $material$) in the CLEVR dataset as follows:

\begin{minted}[bgcolor=\mintedBGColor, tabsize=2]{prolog} 
energise(size(O, C)) :- ... .
energise(material(O, C)) :- ... .
\end{minted}

These energised rules may then be used in the \ASPenergised{} program to infer object properties and relations, by assigning one property per category to each object and define reasoning rules over them:


\begin{minted}[bgcolor=\mintedBGColor, tabsize=2]{prolog} 
1{size(O,Z):size(Z)}1 :- ... .
1{material(O,M):material(M)}1 :- ... .
\end{minted}

\begin{minted}[bgcolor=\mintedBGColor, tabsize=2]{prolog} 
exists(obj(O)) :- obj(O), size(O,large), material(O,metal).
\end{minted}

The ancillary $exists(obj(O))$ rule here denotes that there is an object in the scene, and that it has a large size and is made of metal. \ASPenergised{} solving in this case finds the object properties minimising both the energies, i.e., the size and the material energy.

%
%
%
%
%
%
%
%
%
%
%
%
%

\medskip 

\textbf{Energy-Based Semantics over Stable Models.} \quad  We introduce a distinguished predicate $energy(f,\mathbf{a})$, where $f$ is a function symbol identifying an energy factor and $\mathbf{a}$ is an energized atom.  Given a set of stable models $\mathbf{SM}$[$\Pi]$, for any stable model $M \in$ $\mathbf{SM}$[$\Pi$], $\mathcal{E}(M)$ denotes the set of active energy atoms {$\mathbf{a}$} as follows: 
%
%
{\small
$$\mathcal{E}(M) = \{ (f,\mathbf{a}) \mid energy(f,\mathbf{a}) \in M \}$$
}
Each energy symbol $f$ is associated with a parameterized real-valued function $f_\theta : \mathcal{X} \times \mathcal{D}_f \rightarrow \mathbb{R}$, where $\mathcal{X}$ denotes an input space (e.g., images), $\mathcal{D}_f$ is the domain of arguments of $f$, and $\theta$ are learnable parameters. Given an input $x \in \mathcal{X}$, the energy of a stable model $M$ is defined as the sum of its active energy factors:
{\small
$$E_\theta(M \mid x) = \sum_{(f,\mathbf{a}) \in \mathcal{E}(M)} f_\theta(x,\mathbf{a})$$
}
Learning and inference proceed directly on these energies:

\begin{itemize}

\item \textbf{Inference} identifies preferred stable models by minimizing $E_\theta(M \mid x)$ over $M \in$ $\mathbf{SM}$[$\Pi$].

\item \textbf{Learning} encourages correct models (satisfying logical constraints) to have lower energy than invalid or competing configurations, e.g., via a contrastive objective obtained from ASP solving.

\end{itemize}

Under this formulation, logical constraints encoded in $\Pi$ restrict the set of feasible configurations, while the parameterized energy factors $f_\theta$ assign graded preferences among them. This yields a structured energy-based model in which symbolic reasoning and differentiable scoring are integrated to guide learning and inference directly over complex, constrained combinatorial spaces.



\subsection{Inference \& Learning}
\label{sec:inference_learning}

We implement inference and learning in a modular yet tightly integrated manner within ASPEn (Fig \ref{fig:ASPEn}); practically, we build upon Clingo \cite{Gebser2014-Clingo} based ASP solving and pyTorch based neural processing.

\medskip


\textbf{Inference by Energy Minimisation.}\quad
Given the `energisation' mechanism defined (Sec \ref{sec:synt-semantics}),  the full range of ASP-based inference mechanisms remain applicable and available depending on on the level of expressivity required. For our present purposes, lets focus on the case of the built-in \emph{optimization} functionality of the Clingo ASP solver, by extracting all possible energy terms from the ground program and inject energies computed from the corresponding energy based model. Here, inference is formulated as exact maximum a posteriori (MAP) inference over stable models, i.e., given the energy $E_\theta(M \mid x)$ of a stable model $M \in$ $\mathbf{SM}$[$\Pi$], MAP corresponds to solving:

{\small
$$M^* = \arg\min_{M \in \mathbf{SM}[\Pi]} E_\theta(M \mid x).$$
}

i.e., identifying the stable model with minimal total energy.

\smallskip

From a practical perspective, inference is implemented in the following steps:

\smallskip

\textbf{Step 1.} \emph{Grounding.} \quad
   The logic program $\Pi$ is grounded using clingo, producing a finite propositional program whose stable models correspond to feasible configurations under the symbolic constraints.

\smallskip

\textbf{Step 2.} \emph{Energy Evaluation.} \quad
   For each ground energised atom $energise(A)$, the corresponding neural energy function $f_\theta(x,\mathbf{a})$ is evaluated externally and injected into the program as a numeric cost $energy(A, E)$.

\smallskip

\textbf{Step 3.} \emph{Cost-Based Optimization.} \quad
   MAP inference is implemented using Clingo’s built-in optimization mechanism, using the following minimization statement, which applies the energy $E$ as a cost to the stable model $\Pi$ if atom $A$ is a member of the stable model:

\begin{minted}[bgcolor=\mintedBGColor, tabsize=2]{prolog} 
#minimize  { E : A, energy(A ,E) }
\end{minted}




\smallskip

The total cost of the stable model is given by the sum of all applied costs, i.e. this optimization statement correspond to the additive energy $E_\theta(M \mid x)$.
%
%
%
%

\smallskip

Complete optimization is performed over all stable models, such that the resulting solution is the globally optimal stable model with respect to the injected energy values, and inference is exact with respect to the grounded program and the computed energy values. Logical rules in $\Pi$ restrict the search space to feasible stable models, while the energy terms rank these feasible configurations.  As such, inference in \ASPenergised{} integrates symbolic reasoning and differentiable energy evaluation by compiling the energy-based semantics into weighted stable model optimization, enabling exact MAP reasoning over structured, constraint-driven hypothesis spaces.

\medskip

\textbf{Contrastive Divergence based Learning.}\quad
Learning of the energy functions is performed end-to-end using a contrastive free-energy objective. The central principle here is to assign lower energy to stable models that satisfy supervision constraints, and comparatively higher energy to competing models. The contrastive loss compares two aggregated energy quantities over stable models:

\smallskip

\textbf{\textbullet\quad  Positive Free Energy ($F^+$).}
The positive phase aggregates the energies of all stable models that satisfy a set of logical supervision constraints $C$. Concretely, we consider the subset ${M^+ \in SM[\Pi] \mid M^+ \text{ satisfies } C}$, obtained by augmenting the program $\Pi$ with the supervision constraints prior to solving. The positive free energy marginalizes over this subset via a log-sum-exp aggregation of model energies. This encourages stable models consistent with the supervision constraints to attain lower energy.

\smallskip

\textbf{\textbullet\quad  Negative Free Energy ($F^-$).}
The negative phase aggregates over all stable models $M \in SM[\Pi]$, without enforcing supervision constraints. This represents the unconstrained free energy of the models and provides a reference, that is used to contrast stable models consistent with the supervision constraints against it.

\smallskip

The learning objective is the difference between positive and negative free energies. Minimizing this objective decreases the energy of stable models consistent with the supervision constraints relative to unconstrained stable models, i.e., $\mathcal{L} = F^+ - F^-$. Thus, contrastive divergence shapes the energy landscape defined by $E_\theta(M \mid x)$. Logical rules restrict the admissible hypothesis space, while gradient optimization adjusts the neural energy factors such that correct stable models emerge as energy minima within the constrained combinatorial space.


\section{Applied Evaluation}
We perform problem-domain specific applied evaluation of the overall ASPEn methodology at two dinstinct levels: {\small\bfseries(1)} the basic static case with the MNIST dataset \cite{lecun-mnisthandwrittendigit-2010} aimed at covering a minimal setup; and
{\small\bfseries(2)} a dynamic case involving multi-object tracking of moving objects in real-world settings with the MOT benchmark \cite{MOT2019}.

\subsection{Base Case: ``Addition with MNIST''}
\label{sec:MNIST}

We use the MNIST dataset \cite{lecun-mnisthandwrittendigit-2010} to illustrate the complete end-to-end reasoning and learning cycle within \ASPenergised{}.
For this we are considering the ``Addition'' task \cite{deepProgLog} already described in Example 1 (Sec. \ref{sec:synt-semantics}), where the goal is to learn to assign digit values to images of handwritten digits using a weak supervision setup where only the sum of an image pair is provided for training. 

The `Addition' task within the MNIST dataset is aimed at learning digit values from images, exclusively using only the sum of two digit images as supervision.

\medskip

\textbf{Reasoning.}\quad 
%
%
Reasoning in the MNIST ``Addition'' task refers to finding an assignment of digit values to two digit images, such that the sum of the two digit values matches the sum provided during supervision. To model this we provide the two images and the possible values as facts.

\begin{minted}[bgcolor=\mintedBGColor, tabsize=2]{prolog} 
img(img1). img(img2). digit_value(0..9).
\end{minted}

The following choice rule then assigns one of the possible digit values (0-9) to a given image.

\begin{minted}[bgcolor=\mintedBGColor, tabsize=2]{prolog} 
1{digit(X,V) : digit_value(V)}1 :- img(X).
\end{minted}

We energise the atom $digit(X, V)$ to bind the assignment of a value to a digit image to an energy function that takes the arguments of that atom (i.e., image $X$ and value $V$) as inputs and returns an energy.

\begin{minted}[bgcolor=\mintedBGColor, tabsize=2]{prolog} 
energise(digit(X,V)) :- img(X), digit(X,V).
\end{minted}

The digit energies are obtained using the following EBM to implement a classification task, by providing energies for image-value pairs, where lower energy is interpreted as higher likeliehood.

\medskip

\textbf{Energy Network:}\quad As a neural backbone for predicting and learning digit energies, we employ a convolutional neural network operating on the 28×28 grayscale images from the MNIST dataset. The architecture consists of two convolutional layers, each followed by ReLU activations, to extract hierarchical spatial features. A max-pooling layer reduces the spatial resolution, followed by dropout for regularization. The resulting feature maps are flattened and passed through a fully connected layer with ReLU activation and a final linear layer producing ten class-specific logits corresponding to the digit values (0–9). In the energy-based formulation, these logits are interpreted as negative energies associated with each digit value.



\begin{figure}[t]
    \setlength{\tabcolsep}{6pt}
     \centering
     \sffamily
     \begin{subfigure}[c]{0.49\columnwidth}
         \centering
         \includegraphics[width=\textwidth]{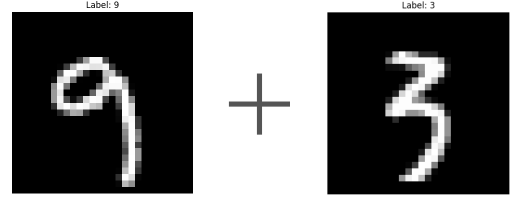}
         \caption{}
         \label{fig:mnist_example}
     \end{subfigure}
     \hspace{0.5cm}
     \begin{subfigure}[c]{0.4\columnwidth}
         \centering
         \begin{tabular}{lcc}
            \toprule \rowcolor{headerblue}
            \textbf{Accuracy} &\textbf{Digit} & \textbf{Sum} \\ 
            \midrule  
	    \cellcolor{headerblue}Single Digit &98.67\%  & 97.54\%  \\ 
	    \cellcolor{headerblue}Double Digit &98.81\%  & 95.33\%  \\ 
            \bottomrule 
         \end{tabular}
         \caption{}
         \label{tab:mnist-addition}
     \end{subfigure}
     
     \caption{\textbf{MNIST Addition}: \textbf{(a)} \emph{Example from MNIST}; \textbf{(b)} \emph{Digit and Sum accuracy}.}
     \label{fig:mnist_combined}
\end{figure}

\medskip

\textbf{Energy based Inference.}\quad 
%
%
At the inference stage, we are searching for the most likely assignment of digit images and values, which translates to minimising the overall energy of the stable model, as described in Sec. \ref{sec:inference_learning}.  
For this we introduce the predicate $addition/1$, which is defined by a rule representing the sum of the two assigned digits:

\begin{minted}[bgcolor=\mintedBGColor, tabsize=2]{prolog} 
addition(N) :- 
	digit(img1,N1), digit(img2,N2), N = N1+N2.
\end{minted}

For instance, for the two digits depicted in Figure \ref{fig:mnist_example} we obtain the following stable model as the best solution, with an overall energy of $-30.35$, obtained from aggregating the individual energies for each digit assignment.

\begin{minted}[bgcolor=\mintedBGColor, tabsize=2]{prolog} 
digit(img1,9).       E: -11.13
digit(img2,3).       E: -19.22
addition(12).
>> Stable Model Energy: -30.35
\end{minted}

%
%
%

%
%
%

\medskip

\textbf{Sum Supervised Learning.}\quad 
We train the network for 6 epochs based on the contrastive divergence principle described in Sec. \ref{sec:inference_learning}. Specifically, we construct the positive and negative free energies ($F^+$ and $F^-$) by aggregating the energies of stable models: $M^+$ accounts for models constrained by the correct sum, while $M$ contrasts this by marginalizing over all unconstrained stable models.
To obtain the different sets of stable models, we are using clingos ability to infuse assumptions while solving, i.e., we add the assumption $addition(Sum) \rightarrow true$, where $Sum$ is given by the added ground truth digit values in the cases for the positive stable models $M^+$ and have no constraints for $M$.


\medskip

\textbf{Results.}\quad 
We evaluate neurosymbolic learning with the MNIST dataset in the single digit, as well as the double digit addition task. 
Digit accuracy and sum accuracy for both tasks, depicted in Table \ref{tab:mnist-addition}, demonstrate that the trained model is capable of correctly recovering digit values from the weakly supervised training examples.
The EBM trained under weak supervision achieves a high per digit classification accuracy of $98.67\%$ respectively $98.81\%$. Crucially, the model successfully scales to weakly supervised two-digit addition, reaching a joint sum accuracy of $95.33\%$. 


\subsection{Case of Combinatorial Reasoning: Visual Question Answering with CLEVR}

CLEVR \cite{Johnson17} is a visual question-answering (VQA) dataset focusing on combinatorial reasoning with image-question pairs. It  has become a standard benchmark for ASP based and neurosymbolic approaches in VQA \cite{EITER_HIGUERA_OETSCH_PRITZ_2022,Yi2018_NSVQA}. The main task in CLEVR is to generate answers to diverse questions regarding the scenes, such as the following:


\begin{quote}
\textbf{Q:} ``\emph{What is the shape of the green metal object that is the same size as the blue object?}''
\end{quote}

Question are given in a json structure involving the questions, answers to the questions, and full supervision for objects and attributes.

\smallskip

\textbf{Reasoning.}\quad
The ASP reasoning logic is given as follows: First we define the possible values for each of the object properties ($color$, $shape$, $size$, and $material$)

\begin{minted}[bgcolor=\mintedBGColor, tabsize=2]{prolog} 
color(red;blue;green;brown;gray;
	purple;cyan;yellow).
shape(cube;sphere;cylinder).
size(small;large).
material(rubber;metal).
\end{minted}

We use choice rules to assign each object exactly one value for each property:

\begin{minted}[bgcolor=\mintedBGColor, tabsize=2]{prolog} 
1{ color(O,C) : color(C) }1 :- obj(O).
1{ shape(O,S) : shape(S) }1 :- obj(O).
1{ size(O,Z)  : size(Z)  }1 :- obj(O).
1{ material(O,M) : material(M) }1 :- obj(O).
\end{minted}

Furthermore, we add spatial relations (\footnotesize{$left\_of$, $right\_of$, $front\_of$}, and \footnotesize{$behind\_of$}) based on the objects 2D positions to handle spatial questions as per the CLEVR specification. 
%
%
%
%
%
Similar as in the MNIST case above, we then energise the property atoms to obtain an energy for the assignment of a specific property to an object.

\begin{minted}[bgcolor=\mintedBGColor, tabsize=2]{prolog} 
energise(color(O, C)) :- obj(O), color(C).
energise(shape(O, S)) :- obj(O), shape(S).
energise(size(O, S)) :- obj(O), size(S).
energise(material(O,M)):-obj(O),material(M).
\end{minted}

These energies then serve as the basis to find the most likely assignment, i.e., to one minimising the overall energy of the resulting stable models.

\smallskip

\textbf{Object Detection and Energy based Attribute Prediction.}\quad 
Object detection is following the approach proposed in \cite{EITER_HIGUERA_OETSCH_PRITZ_2022}, using a pretrained YOLOv3 architecture, fine-tuned for CLEVR for object detection. In addition to object detection, the network also serves as a feature extractor to obtain object-level features for each detected object resulting in a $15 \times 15 \times 303$ feature map per object.
Attribute prediction is based on the extracted feature maps, using EBMs where each attribute type is handled by a specialized neural network that assigns energy values to possible attribute classes. The architecture consists of four attribute-specific heads, each implemented as a multi-layer perceptron (MLP):
The \emph{size} prediction EBM takes as input the concatenation of the visual feature vector and four normalized bounding box coordinates. The network consists of two hidden layers with 256 units each, employing layer normalization and ReLU activations. The output layer produces two energy values corresponding to small and large.
The \emph{color} prediction EBM takes only the visual feature vector as an input. The network consists of three hidden layers using 512 units in the first two layers and 256 units in the third, again with layer normalization, ReLU, and dropout. The output provides energy values corresponding to the eight color classes.
The \emph{material} and \emph{shape} prediction EBMs follow the standard two-layer design with 256 hidden units in each layer, taking only visual features as input. The material head outputs two energy values for rubber and metal, while the shape head produces three values for cube, sphere, and cylinder.
Training of the attribute EBMs is performed fully supervised, using a margin-based hinge loss that encourages correct predictions to have lower energy than incorrect ones by at least a margin $m$.

\begin{figure}[t]
     \centering
     \begin{subfigure}[c]{0.5\columnwidth}
         \centering
         \includegraphics[width=\columnwidth]{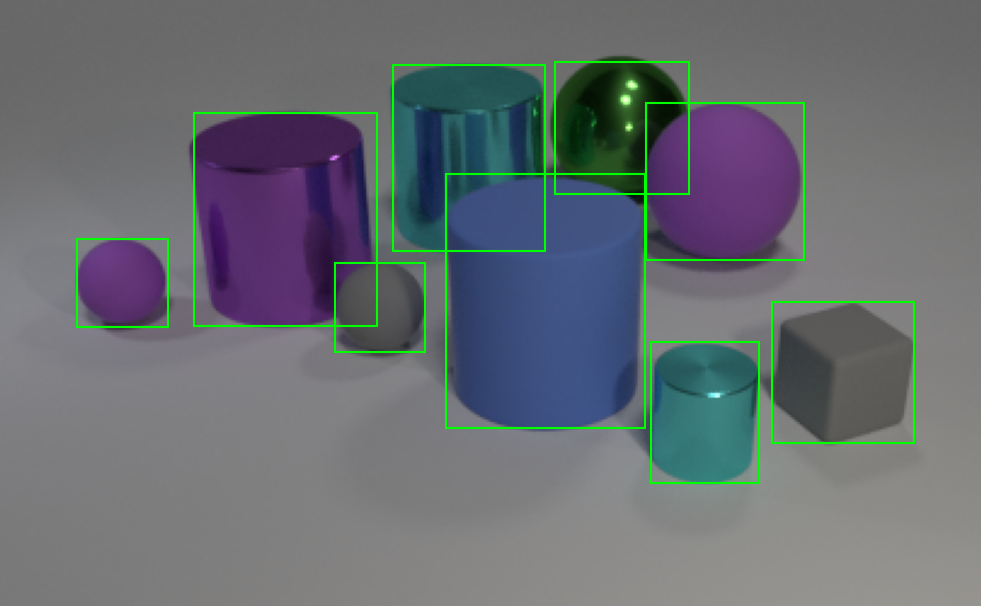}
         \caption{}
         \label{fig:clevr_example}
     \end{subfigure}
     \hfill
     \begin{subfigure}[c]{0.4\columnwidth}
    \setlength{\tabcolsep}{6pt}
        \centering
        \begin{tabular}{lc}
	    \toprule
	    \rowcolor{headerblue}
	    \textbf{Attribute}  & \textbf{Accuracy (\%)} \\
	    \midrule
	    \cellcolor{headerblue}{Size} & 99.5 \\
	    \cellcolor{headerblue}{Color} & 71.8 \\
	    \cellcolor{headerblue}{Material} & 86.3 \\
	    \cellcolor{headerblue}{Shape} &  86.2 \\
	    \midrule
	    \cellcolor{headerblue}\textbf{VQA} & \textbf{62.85} \\
	    \bottomrule
	\end{tabular}
        \caption{}
        \label{tab:vqa_results}
    \end{subfigure}
    
     \caption{\textbf{Clevr Dataset}: \textbf{(a)} \emph{Example Scene from CLEVR Dataset}; \textbf{(b)} \emph{Individual attribute accuracy and VQA accuracy on 15,000 CLEVR image-question pairs.}}
     \label{fig:mot_combined}
\end{figure}

\smallskip

\textbf{Energy based Inference for VQA.}\quad 
For solving the CLEVR VQA task, we are using the functional programs representing the questions, which are provided with the dataset, and translate them to ASP rules by parsing the json structure and extracting the individual steps to answer the question, resulting in a stepwise reasoning strategy, such as the following one:






\begin{minted}[bgcolor=\mintedBGColor, tabsize=2]{prolog} 
set_0(O) :- obj(O).
set_1(O) :- set_0(O), color(O,green).
set_2(O) :- set_1(O), material(O,metal).
set_3(O) :- obj(O).
set_4(O) :- set_3(O), color(O, blue).
set_5(O2) :- set_2(O1), set_4(O), 
	size(O1,V), size(O2,V), O1 != O2.
answer(C) :- set_5(O), shape(O, C).
\end{minted}

%
%

This extracted ASP program representing the question is then combined with the logical program above, to generate the answer, i.e., in this case one of the shape attributes. Energy based inference then results in the following stable model and corresponding energies, with the answer to the above question given by the atom $answer(sphere)$.

\begin{minted}[bgcolor=\mintedBGColor, tabsize=2]{prolog} 
color(obj5, green).    E: 22.39
material(obj5, metal). E: -1.47
size(obj5, large).     E: -1.36
shape(obj5, sphere).   E: 0.94
...
color(obj6, blue).     E: 13.46
size(obj6, large).     E: -1.36
...
answer(sphere). 
>>Stable Model Energy: 137.11
\end{minted}

The stable model grounds the complex referring expression (green metal object) to $obj5$, verifies size and match with the blue object ($obj6$), to predict the shape attribute.

\smallskip

\textbf{Results.}\quad 
%
%
%
Overall VQA accuracy of 62.85\% demonstrates that the energy-based approach successfully bridges visual perception and logical reasoning. The energy formulation provides a principled mechanism for visual predictions, enabling the ASP solver to reason over alternative attribute assignments when visual evidence proves ambiguous. 


\subsection{Case of Dynamics: Multi-Object Tracking (MOT) Benchmark}
Multi-object tracking (MOT) based on the ``tracking-by-detection'' paradigm involves associating detections with object tracks over time.\footnote{{\bfseries MOT} is a  computer vision community established benchmark focussing on Multi-Object Tracking (MOT) \cite{MOT2019}., \href{https://motchallenge.net/data/MOT17}{https://motchallenge.net/data/MOT17}}
 The core challenge is to maintain consistent identities across frames under real-world challenges, including occlusion, appearance variation, crowding, and partial observability. In the context of MOT, we first train the EBM separately with extracted features from the YOLO backbone, to associate pairs of detections across frames. Then, during evaluation, we leverage the trained EBM to compute pairwise association energies between active tracks and new detections, populating a structured energy landscape over which the ASP solver reasons.

\medskip

\textbf{ASP-Guided Association with Energy Minimisation.}\quad
Given a set of active tracks and new detections, the solver must find a globally consistent assignment that respects hard structural and geometric constraints, while minimising the total association energy supplied by the EBM. First, tracks and detections are introduced as domain objects. IoU values between each track--detection pair are computed externally and grounded as facts before solving:

\begin{minted}[bgcolor=\mintedBGColor, tabsize=2]{prolog} 
trk(Trk) :- track_id(Trk).
det(Det) :- det_id(Det).
iou(Trk, Det, Iou).
\end{minted}

Then we enforce a \emph{one-to-one} assignment between tracks and detections. Every track must either be matched to exactly one detection or declared halted; symmetrically, every detection must either be matched to an existing track or initialised as a new one:

\begin{minted}[bgcolor=\mintedBGColor, tabsize=2]{prolog} 
{ match(Det, Trk) : det(Det); halt(Trk) } = 1 :- trk(Trk).
{ match(Det, Trk) : trk(Trk); new_track(Det) } = 1 :- det(Det).
\end{minted}

 For each detection matched to a track, we generate an assign predicate, which serves as a control token, causing the assignment in the tracking cycle.

\begin{minted}[bgcolor=\mintedBGColor, tabsize=2]{prolog} 
assign(Det, Trk) :- match(Det, Trk).
\end{minted}

Spatially implausible associations are ruled out via a hard constraint. Any candidate match whose IoU falls below a fixed threshold is forbidden, acting as a symbolic spatial guard independent of appearance energy:

\begin{minted}[bgcolor=\mintedBGColor, tabsize=2]{prolog} 
:- match(Det, Trk), iou(Trk, Det, IOU), IOU < iou_threshold.
\end{minted}




We energise the atom $match(Det, Trk)$, coupling it with an energy function corresponding to the likelihood that detection $Det$ is matched with track $Trk$.

\begin{minted}[bgcolor=\mintedBGColor, tabsize=2]{prolog} 
energy(match(Det, Trk)) :- match(Det, Trk).
\end{minted}

\ASPenergised{} then optimises for the stable model with minimum aggregate EBM energy among all those satisfying above constraints. In essence, the combinatorial structure of valid assignments is expressed declaratively, while the preference among them is determined by the learned energy landscape.

\medskip

\begin{figure}[t]
     \centering
     \begin{subfigure}[c]{0.7\columnwidth}
         \centering
         \includegraphics[width=\textwidth]{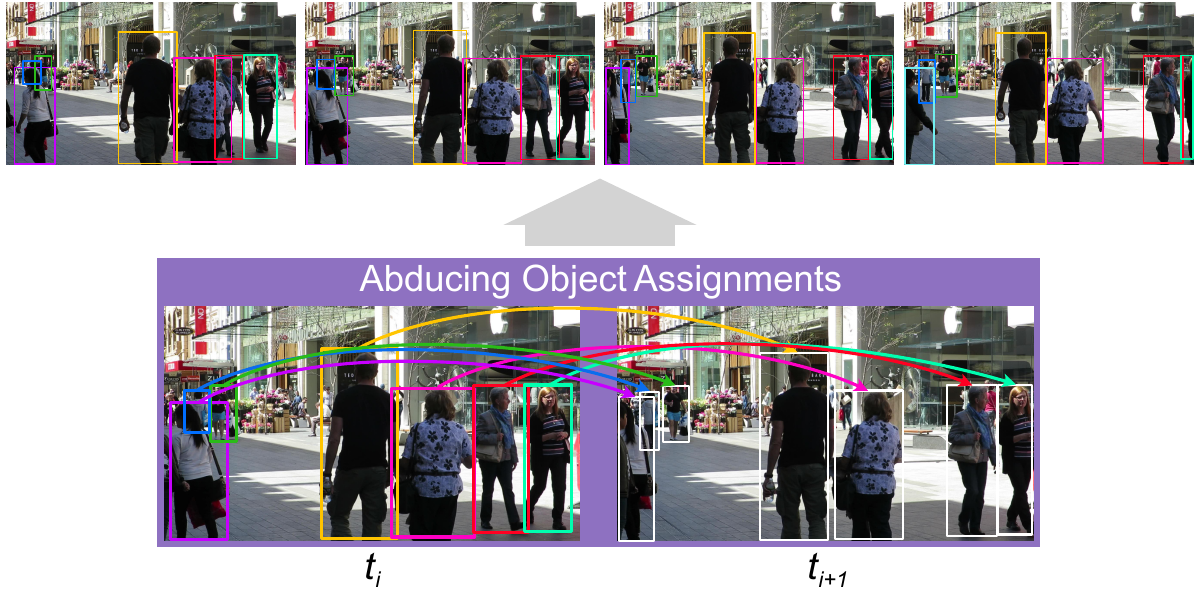}
         \caption{}
         \label{fig:mot_example}
     \end{subfigure}
     \hfill
     \begin{subfigure}[c]{0.25\columnwidth}
    \setlength{\tabcolsep}{6pt}
        \centering
        \begin{tabular}{lc}
           \toprule
           \rowcolor{headerblue}
           \textbf{Metric} & \textbf{Result} \\
           \midrule
            \cellcolor{headerblue}\textbf{HOTA} & 43.26 \\
            \cellcolor{headerblue}\textbf{MOTA} & 36.37 \\
            \cellcolor{headerblue}\textbf{IDF1} & 49.89 \\
           \bottomrule
        \end{tabular}
        \caption{}
        \label{tab:mot-eval}
    \end{subfigure}
    
     \caption{\textbf{Multi-Object Tracking (MOT)}: \textbf{(a)} \emph{Abductive inference for maintaining object tracks}; \textbf{(b)} \emph{Tracking performance with standard MOT metrics}.}
     \label{fig:mot_combined}
\end{figure}

\textbf{Tracking.}\quad
%
The above \ASPenergised{} program is solved for each time-step, providing the optimal assignments of detections to tracks, given the energy values for possible matches. As an example consider, the two frames in Fig. \ref{fig:mot_example}, where on the left are object tracks for $t_i$ and on the right are detections for $t_{i+1}$. Associations in the obtained stable model match these detections to the existing object tracks. 
The tracking logic implements the assignments and maintains the object tracks. In addition, the tracker maintains explicit lifecycle states for each track (\texttt{tentative}, \texttt{confirmed}, or \texttt{deleted}), and coordinates instantiation and deletion of tracks based on thresholds for consecutive matches resp. misses of maintained object tracks.

\medskip

\textbf{EBM Architecture and Training.}\quad
Features are extracted from a pre-trained YOLO26X backbone via ROI-pooled global average pooling over the penultimate feature map, yielding a feature vector for each detection alongside its normalised bounding box. 
The EBM combines two branches, (i) an \emph{appearance branch} processing the pairwise feature maps for obtaining energies corresponding to visual similarity through an MLP; and (ii) a \emph{geometry branch} processing spatial features obtained from the two bounding boxes through a separate lightweight MLP for energies corresponding to geometric similarity. Both outcomes are combined via a learned scalar weight. 
The model is trained with a contrastive loss, which drives the energy of positive pairs below that of all negatives. Training data is constructed from MOT17, where for each anchor detection in frame $t$, a positive is sampled from the same track in a future frame $t{+}\delta$ (with $\delta \leq 30$), and $K$ negatives are drawn from other tracks in the same frame, with hard-negative mining retaining the most energetically confusing negatives during training.


\begin{table}[t]
\centering
\footnotesize
\sffamily
\caption{\textbf{ASPEn for MOT}. \emph{Per-frame Runtime Statistics}.}
\label{tab:runtime_exact}
\setlength{\tabcolsep}{11pt}
\renewcommand{\arraystretch}{1.1}
\begin{tabular}{>{\columncolor{headerblue}}l!{\color{headerblue}\vrule}cccc}

            \toprule
\rowcolor{headerblue}
\textbf{Component} & \textbf{Mean} & \textbf{Std} & \textbf{Min} & \textbf{Max} \\
            \midrule
Feature Extraction Time
& 0.0780
& 0.0062
& 0.0547
& 0.8763 \\

EBM Time
& 0.0101
& 0.0029
& 0.0013
& 0.0693 \\
            \midrule
Grounding Time (Base)
& 0.0013
& 0.0004
& 0.0003
& 0.0046 \\

Grounding Time (Energy)
& 0.0005
& 0.0002
& 8.297e-05
& 0.0031 \\

Total Grounding Time
& 0.0018
& 0.0006
& 0.0004
& 0.0055 \\

            \midrule
Solving Time
& 0.0031
& 0.0011
& 0.0004
& 0.0127 \\
            \bottomrule
\end{tabular}
\end{table}

\begin{table}[t]
\centering\sffamily
\footnotesize
\caption{\textbf{ASPEn for MOT}. \emph{Structural Properties of Computed Stable Models}.}
\label{tab:structure_exact}
\setlength{\tabcolsep}{9pt}
\renewcommand{\arraystretch}{1.2}
\begin{tabular}{>{\columncolor{labelblue}}lcccc}
            \toprule
\rowcolor{headerblue}
\textbf{Average over MOT Sequences:} & \textbf{Mean} & \textbf{Std} & \textbf{Min} & \textbf{Max} \\
            \midrule
Number of Frames per Sequence
& 379.8571
& --
& --
& -- \\

Time per Frame
& 0.2
& --
& --
& -- \\
            \midrule

Number of Energy Atoms per Frame
& 339.3938
& 145.4837
& 30.0
& 1058.0 \\

Number of Model Atoms per Frame
& 36.7236
& 8.0008
& 14.0
& 69.0 \\
            \bottomrule
\end{tabular}
\end{table}




\medskip

\textbf{Results.}\quad
We evaluate on the MOT17 benchmark with HOTA, MOTA, IDF1 as primary metrics (Table~\ref{tab:mot-eval}; training data sequences split into halves for training and validation). Results demonstrate that EBM, trained purely on pairwise contrastive objectives, can serve as an effective matching criteria within the declarative tracking pipeline. HOTA score reflects a balanced account of detection and association quality, while IDF1 captures the consistency of track identities over time; both are directly influenced by the quality of the ASP-guided assignment. The overall results are comparable with trackers of similar character and capacity.
%
%
%
Tables \ref{tab:runtime_exact} and \ref{tab:structure_exact} report verbatim runtime and structural statistics collected during experimental evaluation. These results showcase that ASP with energised atoms scale to real world domains with over 1000 energy atoms per frame and a grounding time for the energy atoms of less then 30\% of total grounding time.


\medskip

\section{Discussion and Related Work}\label{sec:disc}

Systematic neurosymbolic integrations involving answer set programming primarily follow two different mechanisms:  {\small\textbf{(1)}}. One approach involves extensions of ASP introducing \emph{neural predicates} whose truth values are derived from subsymbolic perception models \cite{Yang2020_NeurASP}; and {\small\textbf{(2)}}.  Another approach has been neurosymbolic integration following a hybrid architecture (without neural predicates) in the manner of \cite{outofsight-ijcai2019}, by the development of specialised extensions building upon functional generalisation of ASP, called answer set programming modulo theories (ASPMT) \cite{functionalASP-2013}. 
{In the second approach} building on functional ASPMT, specialised extensions for handling spatio-temporal dynamics in a foundational manner have been developed with ASPMT($\mathcal{QS}$) \cite{TPLP-ASPMTQS}; this provides a unified mechanism for neurosymbolic reasoning about space, events, actions, and change in real-world (real-time) domains involving quantitative, dynamic sensor data such as video \cite{outofsight-ijcai2019,AIJ2021-Suchan,Suchan2025_KR}. Such a foundation then directly powers modular integration with deep learning based computer vision techniques and neurosymbolic capabilities --e.g., encompassing question-answering, relational learning, visuospatial abduction-- driven by  non-monotonic reasoning over continuous space-time structures, and supporting expressive modelling of motion, interaction, and spatio-temporal histories (while also retaining ASP’s model-theoretic foundations).
ASPEn is a novel methodology for declarative neurosymbolic reasoning and learning in a manner that is end-to-end, and designed for challenging real-world dynamic domains. By tightly integrating ASP’s non-monotonic semantics with continuous optimisation, ASPEn offers a principled path toward expressive and explainable reasoning systems especially for dynamic interactive domains. ASP plays a central role in the proposed neurosymbolic methodology; its stable-model semantics provides a robust declarative mechanism for non-monotonic model construction, supporting defaults, exceptions, and abductive explanations that are \emph{indispensable in real-world dynamic domains} (e.g., cognitive robotics, autonomous vehicles). These features fundamentally distinguish ASPEn from logic programming and monotonic probabilistic frameworks, which do not support non-monotonic revision and optimisation over alternative world models.
%
%

\section{Conclusion and Outlook}\label{sec:outlook}

{ASPEn} can be either used as an independent \emph{neurosymbolic reasoning and learning engine} for standalone application purposes, or it may be used as a \emph{experimental platform} to advance basic works aimed at integration of KR and ML research. Our outlook is driven by the observation that existing mainstream ML evaluation paradigms remain ill-suited for systems that must jointly learn and reason under incomplete, evolving, and structured knowledge. Advancements are needed in developing (KR-centric) neurosymbolic benchmarking methods that explicitly test how learned representations interact with declarative constraints, defaults, and optimisation-based model selection. We posit that general frameworks tightly integrating declarative KR formalisms with trainable substrates (such as EBMs) make it possible to examine questions central to KR, e.g., how symbolic structure shapes learning dynamics, how learning alters admissible world models, and how non-monotonic reasoning behaves under uncertainty.  Without such (KR-centric) methodologies (and benchmarks), progress risks collapsing to purely quantitative performance measures, leaving core KR competencies such as abduction,  induction, explanation, and belief revision largely unexamined in end-to-end systems.

%
%

%

\bibliographystyle{splncs04}


\end{document}